%% file: acl_latex.tex
\pdfoutput=1

\documentclass[11pt]{article}

\usepackage[final]{acl}

\usepackage{times}
\usepackage{latexsym}

\usepackage[T1]{fontenc}

\usepackage[utf8]{inputenc}

\usepackage{microtype}

\usepackage{inconsolata}

\usepackage{graphicx}
\graphicspath{../images/}

\usepackage{booktabs} 
\usepackage{graphicx} 
\usepackage{float} 
\usepackage{caption} 
\usepackage{multirow}

\usepackage{todonotes}
\usepackage{footnote}  
\makesavenoteenv{table}  
\usepackage{threeparttable}  
\usepackage{amsmath}
\usepackage{alltt}
\usepackage{todonotes}

\title{A Zero-Shot Open-Vocabulary Pipeline for Dialogue Understanding}

\author{Abdulfattah Safa\textsuperscript{$1, 2$},~
Gözde Gül Şahin\textsuperscript{$1, 2$}
\\[.3em]
\textsuperscript{1}Computer Engineering Department, Koç University, Istanbul, Turkey\\
\textsuperscript{2} KUIS AI Lab, Istanbul, Turkey \\
{\url{https://gglab-ku.github.io/}}\\
}

\begin{document}
\maketitle
\begin{abstract}
\input{sections/abstract}
\end{abstract}

\section{Introduction}
\input{sections/intro.tex}

\section{Related Work}
\input{sections/relWork.tex}

\section{Methodology}
\input{sections/method.tex}

\section{Experimental Setup}
\input{sections/expSetup.tex}

\section{Experiments and Results}
\label{sec:results}
\input{sections/results.tex}

\section{Discussion \& Analysis}
\input{sections/analysis.tex}

\section{Conclusion}
\input{sections/conclusion.tex}

\section*{Ethics Statement}
\input{sections/ethics}
\section*{Limitations}
\input{sections/limitations}

\section*{Acknowledgements}
    This work has been supported by the Scientific and Technological Research Council of Türkiye~(TÜBİTAK) as part of the project ``Automatic Learning of Procedural Language from Natural Language Instructions for Intelligent Assistance'' with the number 121C132. We also gratefully acknowledge KUIS AI Lab for providing computational support.
\bibliography{anthology,custom}
\input{sections/appendix}

\end{document}

%% file: sections/abstract.tex
Dialogue State Tracking (DST) is crucial for understanding user needs and executing appropriate system actions in task-oriented dialogues. Majority of existing DST methods are designed to work within predefined ontologies and assume the availability of gold domain labels, struggling with adapting to new slots values. While Large Language Models (LLMs)-based systems show promising zero-shot DST performance, they either require extensive computational resources or they underperform existing fully-trained systems, limiting their practicality. To address these limitations, we propose a zero-shot, open-vocabulary system that integrates domain classification and DST in a single pipeline. Our approach includes reformulating DST as a question-answering task for less capable models and employing self-refining prompts for more adaptable ones. Our system does not rely on \textbf{fixed slot values} defined in the ontology allowing the system to adapt dynamically. We compare our approach with existing SOTA, and show that it provides up to 20\% better Joint Goal Accuracy (JGA) over previous methods on datasets like MultiWOZ 2.1, with up to 90\% fewer requests to the LLM API. The source code is provided for reproducibility\footnote{\url{https://github.com/GGLAB-KU/open-vocab-dialogue-understanding}}

%% file: sections/intro.tex
    
    Dialogue state tracking (DST) is a critical component of task-oriented dialogue systems, designed to extract and maintain users' goals throughout a conversation~\cite{young-etal-2007-hidden}. The challenge of DST lies in the infinite possibilities of user/agent conversations and the constant evolution of services, schemes, and APIs that dialogue systems interface with~\cite{ren-etal-2018-towards}. While traditional approaches demonstrate reasonable performance within predefined ontologies~\cite{mrksic-etal-2017-neural, liu17c_interspeech}, current research is exploring various strategies for domain transfer. These strategies include adaptation to unseen domains~\cite{li-etal-2021-zero, aksu-etal-2023-prompter}, leveraging non-dialogue QA data to enhance generalization~\cite{liu17c_interspeech}, and framing DST as a question-answering problem using natural language descriptions to enable zero-shot transfer~\cite{lin-etal-2021-leveraging}. However, these approaches still require training on seen domains and closely adhere to domain ontologies.

    A new generation of large language models~(LLMs) such as GPT-4~\cite{openai2023gpt4}, Llama 2 \cite{touvron2023llama} and Gemini 1.0~\cite{geminiteam2024gemini} promise the ability to solve tasks without task-specific fine-tuning, relying instead on the extensive world knowledge acquired from training on vast amount of data. These LLMs have shown remarkable capabilities in in-context learning~(ICL), where the model generates responses based on a natural language prompt and a few examples, achieving significant advancements over fine-tuned methods in few-shot scenarios. Researchers have begun to apply LLMs with ICL techniques to address the DST challenge~\cite{heck-etal-2023-chatgpt, pan2023preliminary, feng-etal-2023-towards}, yet they have not surpassed state-of-the-art~(SOTA) supervised methods, or lacked practicality in terms of number of queries to be executed for \textit{every single turn}, and are highly dependent on the fixed ontology. Furthermore, the majority of these works~\cite{heck-etal-2023-chatgpt, feng-etal-2023-towards} use gold domain labels, usually skipping the domain classification phase, which is nontrivial.\todo[disable, inline, size=\small, color=yellow]{majority (or all): majority}
    
    To address these challenges, we introduce a zero-shot, resource-efficient, and \textbf{open-vocabulary} pipeline system for task-oriented dialogue understanding. Our pipeline starts with domain classification, a crucial phase often overlooked in existing approaches, followed by two complementary approaches for DST. First, we propose DST-as-QA, which transforms DST into a multiple-choice QA problem~\cite{lin-etal-2021-zero}, providing a strong adaptation for smaller or less capable LLMs. Second, inspired by the recent success of self-refining/correcting prompts~\cite{tandon-etal-2022-learning, madaan2024self}, we propose \textbf{DST-as-SRP}---for the first time in literature---that considers LLM as a black-box Dialogue State Tracker, and cast the problem into a well-structured prompt, i.e., Self-Refined Prompt~(SRP). Unlike ontology-based approaches that need to process all possible \textbf{slot value pairs}~\cite{ye-etal-2022-multiwoz} within the ontology, open-vocabulary approaches only use the generic slot definition (which sometimes is common knowledge, e.g., hotel name, restaurant area) and generate/extract the values directly from the \textbf{dialogue}. We then perform a series of experiments with the most recent open-source~(e.g., Llama3) and proprietary language models~(e.g., GPT-4-Turbo) on common DST datasets such as MultiWOZ and SGD~(see \S\ref{sec:results}). We further measure the effect of having access to gold domains~(i.e., with/without domain classification), and ontology; and compare our approaches with various SOTA methods~(both fully trained and zero-shot) on fair settings~(e.g., by using the same LLM). We show that DST-as-SRP achieves new state-of-the-art results with up to \textbf{90\% fewer requests} to the LLM API(compared to a previous SOTA that prompts LLMs for each slot~\cite{feng-etal-2023-towards}), improving the strict Joint Goal Accuracy (JGA) score by 20\%, 3\%, and 2\% on the MultiWOZ 2.1, MultiWOZ 2.4, and SGD datasets, respectively, while still being constrained to \textbf{open-vocabulary} and zero-shot settings unlike current SOTA methods.\footnote{When removing the open-vocabulary constraint for a more fair comparison with current SOTA, we observe an additional 2\% JGA gain for the MultiWOZ datasets as expected.} Finally, we show that the improvements are not simply due to \textit{using a larger language model} but the DST-as-SRP technique in a controlled setup.

%% file: sections/relWork.tex
\paragraph{Dialogue State Tracking as a Question-Answering Problem} The field has seen various approaches to addressing dialogue state tracking (DST) by framing it within a question-answering (QA) context. \citet{gao-etal-2019-dialog} introduced slot-filling as sequential QA tasks, employing a Recurrent Neural Network (RNN) for generating responses. Following that trend, \citet{Tavares2023learning}  fine-tuned a T5 model for the sequential QA tasks and performed zero-shot tests in unseen domains. Similarly, \citet{li-etal-2021-zero} employed manually created questions and a GPT-2 decoder for generating slot values under a supervised framework, then tested for zero-shot applicability. \citet{cho-etal-2023-continual} used a retrieval model to find relevant QA pairs from previous dialogues, then finetuned a T5 model with these samples to adapt to unseen domains. To the best of our knowledge, there exists no in-context learning approach that formulates DST as a QA task. All the aforementioned approaches require a form of fine-tuning and domain adaptation technique. Furthermore, majority of the models extensively depend on existing ontologies for generating answers~\cite{zhou2019multi, cho-etal-2023-continual}, and several others~\cite{Tavares2023learning, li-etal-2021-zero} struggle with the efficiency, since they need to generate tremendous amount of questions per turn (see \S \ref{ssec:efficiency}). 

\paragraph{LLMs Zero-Shot Dialogue State Tracking} 

\citet{pan2023preliminary} was the first to explore ChatGPT's zero-shot dialogue understanding capabilities using schema-based prompts, achieving notable success in basic slot-filling tasks but encountering issues with multi-turn dialogues. Then, \citet{heck-etal-2023-chatgpt} assessed ChatGPT's performance across various slot types, employing custom prompts for different interaction lengths and slot types, slightly lagging behind state-of-the-art zero-shot models in handling complex slots. After that, \citet{feng-etal-2023-towards} attempted to track the slot values of the dialogue turns one by one, appending all possible slot values from schema and outperformed the current zero-shot models. Yet, this approach needs extensive number of queries for every turn, and unable to handle the \texttt{dontcare} slot values. Despite the remarkable results reported by these models, their dependence on predefined schemas and ontologies limits their practical utility in DST, particularly in environments where new entities, types, and services are continuously introduced.
    
While both QA and DST-as-zero-shot approaches have shown promise, our research specifically addresses the challenge of identifying relevant slots to query \textit{efficiently}, and \textit{generating possible answer options} for the questions (open-vocabulary), instead of using ontology-dependent answer templates. 
Furthermore, we experiment with a realistic end-to-end pipeline that includes domain classification, unlike approaches that treat domains as ``given'' and use the gold annotation.

%% file: sections/method.tex
\begin{figure*}[!htbp]
    \centering
    \includegraphics[width=\textwidth]{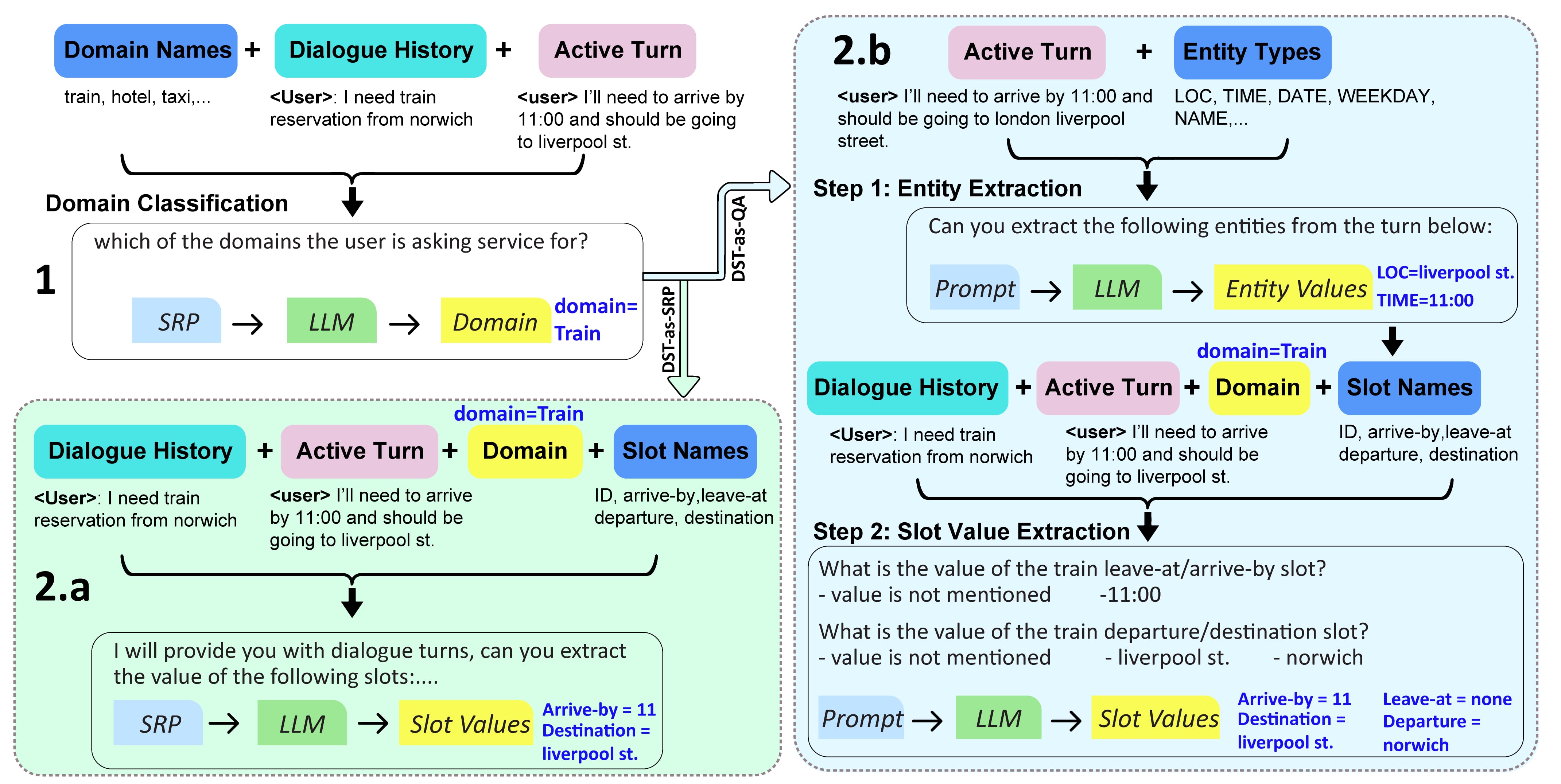}
   \caption{Overview of the architecture, comprising two stages: \textbf{1. Domain Classification} and \textbf{2. Dialogue State Tracking (DST)}. DST can be performed via either \textbf{2.a: DST-as-SRP} or \textbf{2.b: DST-as-QA}. The color scheme is as follows: prompts have a cyan background, schema has a blue background, results are in blue, and output stages have a yellow background. (The prompts in this figure are illustrative. For actual prompts, please refer to App.~\ref{app:prompt-template})}
    \label{fig:arch}
\end{figure*}

An overview of our methodology is given in Fig.~\ref{fig:arch}. Our pipeline begins by identifying the active domain for each turn. For domain classification, the active turn and all preceding turns in the dialogue history are passed to the model, which determines the domain using a specific prompt tailored to the language model~(see \S~\ref{ssec:domain_classification}). Next, we extract the values of the selected domain slots using two approaches: question answering~(see \S~\ref{ref:ssec:DST-QA}) and self-refined prompting~(see \S~\ref{ref:ssec:DST-SRP}).

\subsection{Self-Refined Prompt}
\label{ssec:self_correct}
    Self-Refined prompt~(SRP) approach \cite{madaan2024self} iteratively refines the prompts harnessing the adaptive capabilities of language models (LMs). It has been shown to benefit many NLP tasks such as Code Optimization, Sentiment Reversal and mathematical reasoning that encouraged us to employ it both for domain classification and dialogue state tracking.
    Initially, a basic prompt template \( P \) serves as either a simple task description outlining what the LM needs to accomplish or as a preliminary version of the prompt that will be iteratively improved. The LM generates an initial output based solely on \( P \), without any additional context or task-specific information. Following this, the prompt itself includes instructions for the LM to analyze its output, identifying ambiguities, inaccuracies, or gaps in understanding. The \textbf{SELF-REFINE} approach is then employed, where the LM provides feedback on its own output, such as identifying specific errors or suggesting improvements. The model uses this feedback to refine \( P \), generating a new version \( P' \) that addresses the identified shortcomings, such as including more specific instructions or restructuring the format. This cycle of self-assessment, feedback, and refinement continues until the prompt \( P \) evolves into a version \( P' \) that meets a stopping criterion. The stopping criteria are reached either when the model begins to make only minor changes, suggesting that further refinements would not lead to notable improvements, or when the iteration limit is reached to ensure efficiency. The process ensures that the model optimizes its instructions and task execution autonomously, ultimately resulting in a prompt that consistently achieves better task performance across diverse contexts.
    
    In this study, each LLM uses a set of custom self-refined prompts tailored to its specific characteristics and capabilities. The decision to employ model-specific prompts was driven by the recognition that different LLMs, due to their unique architectures and training data, may respond optimally to slightly different linguistic cues and task formulations~\cite{zhu2024promptrobustevaluatingrobustnesslarge}.

\subsection{Domain Classification}
\label{ssec:domain_classification}
    
    {Understanding user intent and their specific requests is crucial for dialogue state tracking (DST) and begins with identifying the target domains. This pivotal step is often overlooked in DST models, which either rely on a predefined set of domains from the dataset or attempt to indiscriminately track slots across all domains for each interaction. To develop a robust DST pipeline that accommodates multiple domains per turn, we classify the domains for each turn within the dialogue by incorporating dialogue history into the equations. Specifically, the input to the classifier is both the turn and the dialogue history. Let \( D = \{d_1, d_2, \ldots, d_n\} \) be the set of predefined domains and let \( H_{t-1} \) represent the dialogue history up to turn \( t \), with \( U_t \) being the user's utterance at turn \( t \). Then we define the classification function \( f \) as \( f(U_t, H_{t-1}) \to \{d_{i_1}, d_{i_2}, \ldots, d_{i_k}\} \), where \( \{d_{i_1}, d_{i_2}, \ldots, d_{i_k}\} \subseteq D \) represents the set of domains classified for turn \( t \). To approximate this function, we use a multi-label classification approach that can guide the model to infer the appropriate domains from the dialogue. The final prompt template is given in App.~\ref{app:dc_prompt-template}.
    
    As domain classification is the first step in the pipeline, its accuracy is critical for downstream tasks like slot tracking. Any misclassification can have a cascading impact. To mitigate this, we ensure that domain classification is independent for each dialogue turn. For instance, if a turn is classified as taxi-related, we don't carry over the ``taxi'' to subsequent turns. 
    To further improve the accuracy, we enforce strict guidelines in the prompts (in the initial task description provided for the LLM to generate the SRP) to ensure the classification remains grounded in the specific dialogue context and is not biased toward domains that might be inferred from specific phrases. For example, given the utterance ``I want a taxi to go to the hotel.'', the domain should be \textit{strictly} classified as ``taxi'' rather than ``hotel'' and ``taxi''. This is also to solve the classification of the closing turns (e.g., ``have a nice day'', ``anything else for today''), which contain generic terms and hurt the performance as also showed in \citet{hudecek-dusek-2023-large}. 

\subsection{DST as Question Answering}
\label{ref:ssec:DST-QA}

    Our DST-as-QA approach is given in Fig.~\ref{fig:arch}(2.b), DST operates methodically at each user turn, indexed by \(i\), incrementally updating the state as the dialogue progresses. First, we identify the set of entity types from the dataset README files (e.g., TIME, LOCATION etc...). Then, we extract named entities via zero-shot prompting given the user utterance \(U_i\) resulting in a set \(E_i = \{e_1, e_2, \ldots, e_n\} \), which includes all named entities identified during the turn. For instance, if a user says ``I'll need to arrive by 11:00 and it should be going to London Liverpool Street'', the named entities extracted could include ``11:00'' for the TIME entity and ``London Liverpool Street'' for the LOCATION entity. Next, these entities are matched by type to corresponding slots using a predefined matching function given in App. \ref{app:entity-slots}, forming matched pairs \(m_i = \{e_i, \{s_{0..n}\} \}\). Note that the mapping can be one-to-many. For instance, ``11:00'' is matched both to the \texttt{leave-at} and \texttt{arrive-at} (TIME), and ``London Liverpool Street'' is matched to the \texttt{departure} and \texttt{destination} slots (LOCATION).
    Then, for each slot \(s_i\) that has been matched with an entity type (e.g., \texttt{leave-at}), a multiple-choice question with the options: \textit{found entity value} (e.g., 11:00), and None. There are two exceptional cases: First case is detecting a \texttt{dontcare} slot. In that case, it is added to the options. The second case occurs when slots of the same type are captured in previous turns (e.g., booking a hotel and then a taxi to the booked hotel). Then, we add the values of these slots to the options to allow the model to handle cross-referencing issues (i.e., where a slot value in one domain depends on a slot value from another domain). We finally concatenate the question with the active turn \(U_i\), and the dialogue history \(H_{i-1}\) and the options and then prompt the model to select one of the options.
    The dialogue state at index \(i\), denoted \(D_i\), is updated with these selections, continually adapting with each user turn. This process ensures precise tracking and contextual updating of the dialogue, facilitating a dialogue state that dynamically adapts to user inputs and maintains contextual relevance throughout the interaction.

    Here, we create questions for only subsets of the slots that have extracted values of the same type, rather than for all the slots in the schema~\cite{lee-etal-2021-dialogue, lin-etal-2021-leveraging, li-etal-2021-zero}, which should significantly reduce computational costs. To illustrate, consider the MultiWOZ dataset, which includes 61 slots across 8 domains. If we were to generate questions for each slot in all 7372 turns of the test split, this would necessitate a total of 449,692 questions. However, by targeting only relevant subsets, we can dramatically decrease this number. Additionally, our approach does not rely on predetermined slot values from the schema or database \cite{lee-etal-2021-dialogue, feng-etal-2023-towards}; instead, we extract dynamic values directly from user and system turns (open-vocabulary). This strategy allows the dialogue system to adapt more flexibly and accurately to the actual data presented during each interaction, reflecting real-world usage more effectively than static, pre-defined lists could. Furthermore, we utilize multiple-choice questions rather than open-ended ones \cite{li-etal-2021-zero, lin-etal-2021-leveraging, Tavares2023learning}, providing the language model with specific, contextually relevant options. This approach minimizes the risk of misinterpretation and improves the precision of the language model's responses, ensuring that the dialogue management is both accurate and contextually appropriate. The questions prompts are given in App.~\ref{app:entity-extraction-prompt-template} and App.~\ref{app:slot-value-question-prompt-template}.

\subsection{DST as Self-Refined Prompt}
    \label{ref:ssec:DST-SRP}
    The general SRP approach is explained in Sec.\ref{ssec:self_correct}. Here, we explain the structure of the final revised version of the prompt App.~\ref{app:dst-prompt-template}. It is divided into three main sections: task, schema, and regulations. The \textbf{task} section specifies actions for the language model, such as identifying updated or confirmed slots based on user input. The \textbf{schema} section provides a structured framework for the model by listing the slots to be tracked along with their descriptions. Finally, the \textbf{regulations} section defines the precise conditions and expected output formats, ensuring accurate updates of slot values.
    
    Unlike previous approaches ~\cite{hudecek-dusek-2023-large}, our method uses a consistent prompt template with adaptable slot names across all domains, making it flexible and efficient for supporting new slots with minimal modifications. Moreover, we instruct the model to track all slots simultaneously rather than one at a time, in contrast to \cite{feng-etal-2023-towards}, \textit{further improving efficiency}. To illustrate, if we were to generate prompts for each slot individually across all 7372 turns in the MultiWOZ test split, similar to previous methods, this would result in 449,692 prompts (could be long prompt due to the task description length). Our approach, however, reduces this number significantly by consolidating all the domain slots into one comprehensive prompt per turn domain. It's worth noting that our schema avoids listing examples or potential slot values, focusing instead on precise task descriptions. This strategy improves the model's adaptability to varied dialogue contexts. Finally, unlike several previous approaches \cite{feng-etal-2023-towards} that don't distinguish between ``None'' and \texttt{dontcare} slots, our method explicitly handles such cases, preventing misinterpretations and inaccuracies.

%% file: sections/expSetup.tex
  
    \subsection{Datasets}
        
        We conduct experiments using test splits of two most common datasets for multi-domain task-oriented dialogue.
        
        \paragraph{Schema-Guided Dialogue (SGD)} SGD~\cite{rastogi2020towards} is the most challenging dataset, consisting of over 16,000 conversations between a human user and a virtual assistant. It encompasses 26 services across 16 domains, such as events, restaurants, and media. Notably, SGD introduces unseen domains in the test set, challenging the generalization ability of the model.
        
        \paragraph{MultiWOZ}~\cite{budzianowski-etal-2018-multiwoz} has had a significant impact on task-oriented dialogue research, serving as the first substantial public dataset available to researchers in this domain.
        The dataset includes over 10K conversations across eight domains, such as Train, Taxi, Bus, Hotel, Restaurant, Attraction, Police, and Hospital. Following the foundational MultiWOZ 2.0, the dataset underwent several significant annotation fixes and improvements~\cite{eric-etal-2020-multiwoz, zang-etal-2020-multiwoz, han2020multiwoz, ye-etal-2022-multiwoz}. We chose to use versions 2.1 and 2.4 because 2.1 is the mostly widely in literature and 2.4 is the most stable version, and using both allows us to compare our results with previous work in a consistent manner.
 
    \subsection{Evaluation}
        Following the previous works~\cite{hudecek-dusek-2023-large, ye-etal-2022-metaassist, feng-etal-2023-towards}, we use accuracy for the domain classification task, and Joint Goal Accuracy (JGA) as main metric for the DST task. We also report Average Goal Accuracy (AGA) metric for the DST task in one experiment to compare the performance with the baseline model. JGA is the primary metric for DST evaluation and represents the ratio of dialogue turns for which the entire state is correctly predicted\footnote{Following \citet{nekvinda-dusek-2021-shades} and \citet{feng-etal-2023-towards}, we use Fuzzy Match to compare the slots values.}. AGA represents the average accuracy of the active slots in each turn. A slot becomes active if its value is mentioned in the current turn and is not inherited from previous turns.

    \subsection{LLMs}
        We identify 5 popular and capable\footnote{These LLMs are chosen from the models that rank high in the \hyperlink{https://huggingface.co/spaces/lmsys/chatbot-arena-leaderboard}{Hugging Face Chatbot Leaderboard}.} LLMs that are diverse in architecture, scale, and avalability, namely as GPT-4\footnote{Model variant: \texttt{gpt-4-turbo-preview}}~\citep{openai2023gpt4}, Gemini \footnote{Model variant: \texttt{gemini-1.0-pro}}~\citep{geminiteam2024gemini}, LLAMA3\footnote{Model variant: {Meta-Llama-3-70B-Instruct}}~\citep{meta_llama_3}, QWEN \footnote{Model variant: \texttt{qwen1.5-32b-chat}}~\citep{qwen}, and Mixtral\footnote{Model variant: \texttt{Mixtral-8x7B-v0.1}}~\citep{jiang2024mixtral}. We use the respective model APIs, where available. Due to computational constraints, we prioritize the best performing open-source and proprietary models, namely as Llama 3 and GPT-4-Turbo on the SGD dataset, and don't perform experiments with others. Preliminary tests on the subsets were conducted for each model to identify optimal \texttt{temperature} and \texttt{top\_p} parameters. The best-performing configurations were then applied to the entire datasets for a thorough evaluation. Further details on the prompt parameters are given in App.~\ref{app:llms-parameters}.

%% file: sections/results.tex
    \subsection{Domain Classification}
        We present the domain classification accuracy on MultiWOZ and SGD in Table \ref{tab:domain_class}. These results reveal that while all models are effective at domain classification tasks, Gemini demonstrates a slightly better performance in the MultiWOZ datasets. The improved accuracy on MultiWOZ 2.4 suggests that advancements in dataset quality and model improvements contribute to better overall performance. Both Llama3 and GPT-4 perform worse on SGD. We believe this is due to higher number of domains (e.g., 7 in MultiWOZ domains versus 16) and the considerable similarity between the SGD domains (see App.~\ref{app:sgd-cd-analysis}). We observe that GPT-4-Turbo yields consistently high classification accuracy, showing robust performance across schemes.
        \todo[disable,inline, size=\small]{Maybe they expect some examples}
        \todo[disable,inline, size=\small, color=yellow]{There is already detailed analysis in the appendix. I can add some example there}
        \begin{table}[!hbtp]
        \centering
        \scalebox{0.7}{
            \begin{tabular}{lccc}
                \toprule
                \textbf{LLM} & \multicolumn{2}{c}{\textbf{MultiWOZ}} & \textbf{SGD} \\
                \cmidrule{2-3}
                & \textbf{2.1} & \textbf{2.4} & \\
                \midrule
                GPT4-Turbo & 94.56 & 95.98 & \textbf{93.38} \\
                Llama 3 & 93.09 & 94.52 & 85.49 \\
                Qwen 1.5 & 94.01 & 95.37 & - \\
                Gemini 1.0 & \textbf{94.89} & \textbf{96.21} & - \\
                Mixtral v0.1 & 90.49 & 91.72 & - \\
                \bottomrule
            \end{tabular}
        }
        \caption{Domain Classification accuracy varies across different LLMs when applied to the MultiWOZ 2.4 (has 7 domains), and SGD (has 18 domains) datasets}
        \label{tab:domain_class}
        \end{table}

    \subsection{Dialogue State Tracking}
       Our main results for the DST task are given Table~\ref{table:dst_results}, offering a comparative analysis of various language models on the MultiWOZ 2.1, 2.4 and SGD datasets. We provide our end-to-end pipeline results where we use the output from our domain classification module with \texttt{Pred} column. Since previous state-of-the-art methods often assume access to gold domains, we also provide results with the same setting shown with the \texttt{Gold} column for a fair comparison. As expected, we observe a performance drop for the end-to-end case compared to the gold. While the drop is small for the QA approach, the gap is substantially higher for our state-of-the-art performing approach, SRP both with GPT-4-Turbo and Llama3.  We couldn't find a certain explanation for this, but it could be related to QA and domain classification both relying on the LLM's ability to answer questions. Having the same context and asking different questions might lead to similar trends in incorrect answers, resulting in less impact on the overall pipeline. \todo[disable,size=\small, inline]{you need to tell more. We also believe that one reason would be the differences between the level of robustness to noise, i.e., large models are shown to be more robust to syntactic noise}
       \todo[disable,size=\small, inline, color=yellow]{Done. Paragraph updated with citation}
       Additionally, we notice that the performance gap between MultiWOZ 2.1 and 2.4 is smaller for our models compared to fully-trained models. We believe that one reason would be the differences between the level of robustness to noise, i.e., large models are shown to be more robust to syntactic noise~\cite{zheng-saparov-2023-noisy}. This observation is consistent with the zero-shot results reported by \citet{pan2023preliminary}.

       We find that the ranking of the LLMs are similar for both QA and SRP approaches: Gemini/Mixtral, Llama3, and GPT-4-Turbo, mostly showing high correlation with the model size with some exceptions. Except from Mixtral v0.1 and Qwen 1.5\footnote{Qwen 1.5 couldn't distinguish between the task and the dialogue, and provided response to the user's utterance in the dialogue}, all language models yield considerably higher JGA scores with SRP compared to QA. This issue might be due to error cascading in QA, where an incorrect entity value extracted leads to incorrect slot values being selected in subsequent steps. On the other hand, the performance gap between the best and worst performing models are significantly lower for QA compared to SRP---26\% for QA versus up to 41,45\% for SRP. We believe there are two main reasons for this result. First of all, QA method seems particularly effective for smaller models by framing DST as a simpler, question-answering task, which enables these models to focus on specific slots. This reduces the processing overhead and improves efficiency for the smaller language models. Second, SRP requires a deeper understanding of the task, which is only demonstrated by the larger LLMs like GPT-4-turbo and Llama 3. They show higher capacity in following the structured instructions provided by the prompts and incrementally updating slot values accurately. SRP demonstrates the potential for handling dynamic and complex dialogue scenarios by large language models, as it guides the models through structured instructions for efficient zero-shot DST.
        
       The comparison with SOTA models reveals that our approach generally surpasses both zero-shot and fully-trained models. While zero-shot models often exhibit lower performance, even those trained with comprehensive data struggle to match our results. For instance, our SRP approach using the open-source model Llama3 outperforms TOATOD~\cite{bang-etal-2023-task}---a recently introduced fully trained model---by around 7 points. Furthermore, models employing ontologies, such as Preliminary Evaluation of ChatGPT \cite{pan2023preliminary}, LDST \cite{feng-etal-2023-towards}, and TOATOD \cite{bang-etal-2023-task}, also fall behind our SRP with GPT-4-Turbo without any access to ontology  \todo[disable,size=\small, inline]{Please write it in the text AGAIN that the cited works use GOLD domains. SO the comparison is not fair}
        \todo[disable,size=\small, inline, color=yellow]{Added to the text, in addition to the table}values. Note: The majority of the cited SOTA methods use golden domains for evaluation. Comparing these directly to our predicted domain results may not be entirely fair, as it does not reflect equivalent testing conditions. Similar to domain classification, we observe consistently higher scores for MultiWOZ 2.4 from all approaches and models, signaling the increased quality of the dataset.

        \begin{table*}[ht]
            \centering
            \scalebox{0.75}{
            \begin{tabular}{llcccccc}
                \toprule
                \multicolumn{2}{c}{\textbf{Models}} & \multicolumn{2}{c}{\textbf{MultiWOZ 2.1}} & \multicolumn{2}{c}{\textbf{MultiWOZ2.4}} & \multicolumn{2}{c}{\textbf{SGD}} \\
                \cmidrule(lr){3-4} \cmidrule(lr){5-6} \cmidrule(lr){7-8}
                & & \textbf{Gold} & \textbf{Pred} & \textbf{Gold} & \textbf{Pred} & \textbf{Gold} & \textbf{Pred} \\
                \midrule
                \textbf{Full}
                & TOATOD \cite{bang-etal-2023-task}   & 54.97 & - & - & - & - & - \\
                & D3ST \cite{zhao2022descriptiondriven}   & \underline{57.80} & - & 75.90 & - & 86.40 & - \\
                & LDST \cite{feng-etal-2023-towards}      &  56.69 & - & \underline{79.94} & - & 84.47 & - \\
                & paDST \cite{ma2022e2edst}      & - & - & - & - & \underline{86.53} & - \\
                & Schema-Driven Prompt. \cite{lee-etal-2021-dialogue}     & 56.66 & - & - & - & - & - \\
                \midrule
                \textbf{Zero-Shot}
                & Prel. Eval. of ChatGPT \cite{pan2023preliminary}     & -  & \underline{60.28} &  -& 64.23  & - \\
                & LDST \cite{feng-etal-2023-towards}      & - & - & \underline{83.16} & - & \underline{84.81} & - \\
                \midrule
                \textbf{Ours}
                &\textbf{QA}\hspace{2.35em} Gemini 1.0 (1.6T Param.) & 50.21 & 49.73 & 50.90 & 50.28 & - & - \\
                &\hspace{4em} Mixtral v0.1 (45B Param.)& 51.07 & 50.61 & 51.23 & 50.72 & - & -  \\
                &\hspace{4em} Llama 3  (70B Param.) & 56.35 & 56.03 & 57.39 & 57.07 & - & - \\
                &\hspace{4em} Qwen 1.5 (32B Param.) & 58.29 & 57.45 & 59.01 & 58.43 & - & - \\
                &\hspace{4em} GPT-4-Turbo & \underline{75.23} & \underline{74.52} & \underline{76.96} & \underline{76.01} & - & - \\
                \cmidrule(lr){2-8}
                &\textbf{SRP}\hspace{2.3em} Mixtral v0.1 (45B Param.)& 45.51 & 44.59 & 44.85 & 43.81 & - & - \\
                &\hspace{4em} Gemini 1.0 (1.6T Param.)& 68.79 & 65.93 & 69.93 & 67.00 & - & - \\
                &\hspace{4em} Llama 3 (70B Param.)& 70.01 & 65.06 & 71.27 & 66.20 & 76.34 & -\\
                &\hspace{4em} Llama 3+Ontology (70B Param.)& 76.39 & - & 76.73 & - & - & -  \\
                &\hspace{4em} GPT-4-Turbo & 84.02 & \textbf{77.10} & 86.30 & \underline{79.58} & \textbf{88.70} & - \\
                &\hspace{4em} GPT-4-Turbo+Ontology & \textbf{86.01} & - & \textbf{88.33} & - & - & - \\
                &\hspace{4em} Qwen 1.5 (32B Param.)& - & - & - & - & - & -  \\
                \midrule
                
            \end{tabular}
            }
            \caption{Comparative performance of DST models on MultiWOZ and SGD datasets using JGA with ground-truth (\textbf{gold}) and predicted domains (\textbf{pred}). \textbf{Full:} Methods trained on full training split, \textbf{+Ontology:} Access to gold domain ontology, \textbf{Pred:} Using predicted domains, \textbf{Gold:} Using gold domains. ``-'' denotes results being not available. Overall best scores are given in \textbf{bold}, best scores for each category is \underline{underlined}.}
            \label{table:dst_results}
        \end{table*}
       Next, as mentioned previously, for fair comparison with SOTA models, we assess the impact of using slot values from ontology in our approaches. After deriving the slot values, we incorporate them into the slot description when employing self-refined prompts in DST. Due to resource limitations, we only experiment with the best-performing open-source and proprietary LLMs (GPT-4-Turbo and Llama 3). As given Table~\ref{table:dst_results}, incorporating ontology-derived details (when available) into models like GPT-4-Turbo and Llama 3 enhances their ability to accurately fill dialogue slots, resulting with around 2\% and 6\% improvements for GPT-4 and Llama3 respectively. This suggests that even well-performing models can benefit from the integration of more structured data, leading to improvements in how they process and respond within conversational contexts.
        \todo[disable,inline, size=\small]{if it makes sense to add AGA, can you make another table in the appendix with aga scores?}
        \todo[disable,inline, size=\small, color=yellow]{JGA is not available for this case in the SOTA. Also I found that adding AGA will make some confusion to the main table or additional table. Only one or two SOTA reported AGA and sometimes per domain.}
        \begin{table}[ht]
        \centering
        \scalebox{0.63}{
                \begin{tabular}{lccc}
                    \toprule
                    \textbf{Dataset} & \textbf{Model} & \textbf{LLM} & \textbf{AGA*} \\
                    \midrule
                    MultiWOZ 2.2 & LDST (Multi-Return) \tnote{a} & GPT-3.5-Turbo & 81.50 \\
                    MultiWOZ 2.2 & Ours & GPT-3.5-Turbo & 90.98 \\
                    \bottomrule
                \end{tabular}
            }
            \caption{Comparison of LDST and SRP on the MultiWOZ 2.2 Test Split in the same test setup.\\\footnotesize{*We report AGA (not JGA) here to compare with the provided AGA scores in LDST.}}
            \label{tab:prompt-analysis}
        \end{table}
    
       Since the language model version might have a large impact on results, we compare LDST~\cite{feng-etal-2023-towards} and SRP using the same model which is shown to be capable: GPT-3.5-Turbo. By using the same dataset, slot descriptions, possible values, and language model across different prompts (ours versus LDST), we show that SRP outperforms LDST by a large margin (9\%) as shown in Table~\ref{tab:prompt-analysis}.

%% file: sections/analysis.tex
\subsection{Error Analysis}
\label{ssec:domain-error}
    \begin{figure}[!htbp]
        \centering
        \includegraphics[width=0.35\textwidth]{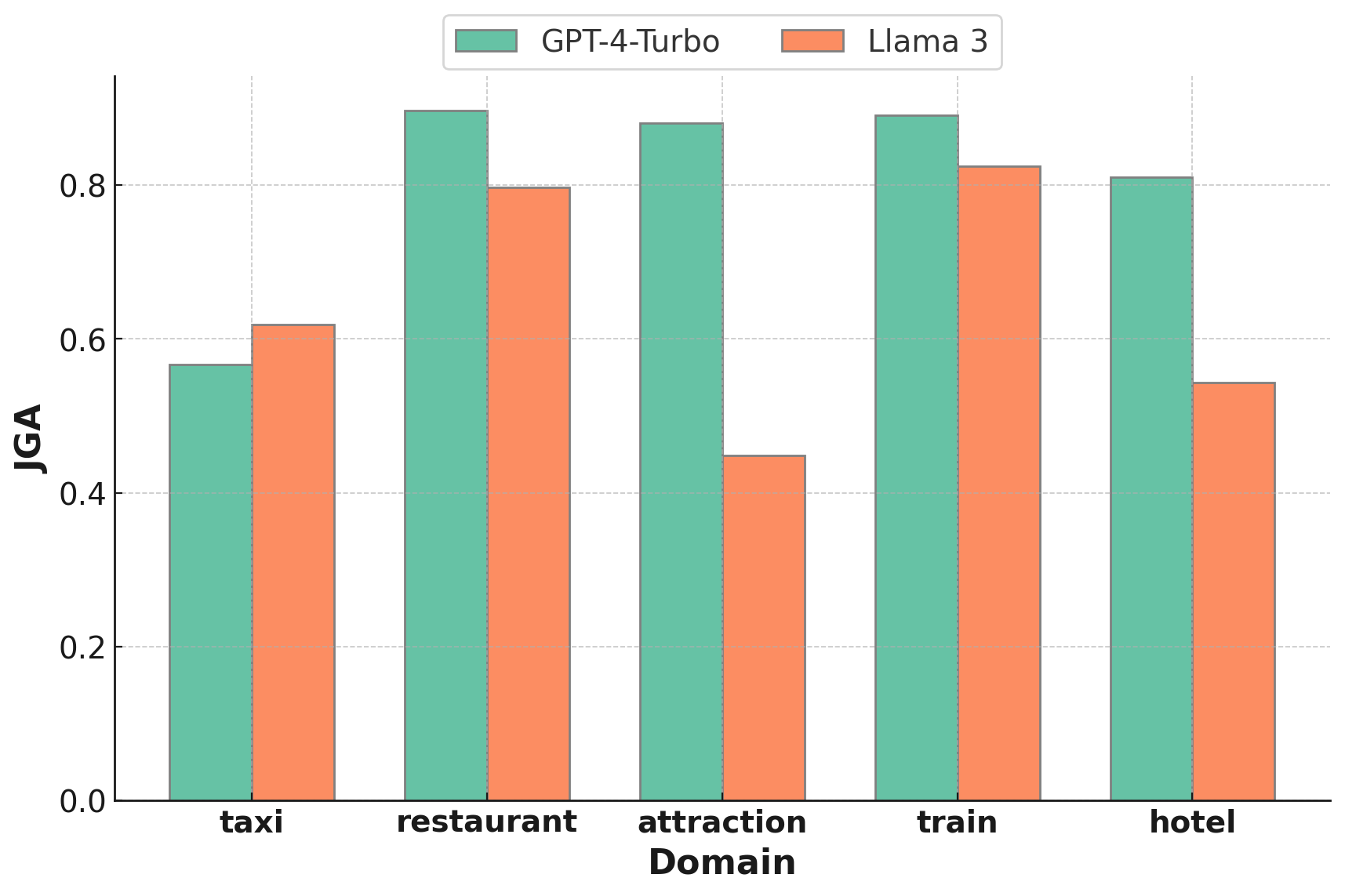}
        \caption{JGA Comparison Across Domains in MultiWOZ 2.4 Dataset for GPT-4-Turbo and Llama 3 Models}
        \label{fig:domain_jga}
    \end{figure}
    
    To get a better picture of the SRP approach performance with DST, we analyze the results of GPT-4-Turbo and Llama 3 per domain. Fig.~\ref{fig:domain_jga} highlights that GPT-4-Turbo exhibits notable performance dips in the taxi domain. This performance issue could be related to the normalized \texttt{leave-at} and \texttt{arrive-by} time values in MultiWOZ 2.4 and the need for extra reasoning to handle cross-referencing slot values. Llama 3, while competent in certain areas, displays more variability. It performs well in the train domain but struggles notably in the attraction and hotel domain. More detailed analysis per slot can be found in App.~\ref{app:mwz-srp-slot-jga}.
\subsection{Analysis of LLM Prompt Requests}
\label{ssec:efficiency}
   Finally, we evaluate the efficiency and scalability of our models by analyzing the number of LLM prompt requests required, a key metric for computational cost and deployment viability. Table~\ref{tab:llm_queries} shows the average API calls per dialogue (see App.~\ref{app:llm-requests} for details on LLM request metrics and calculation methods) for MultiWOZ 2.4. The results show that our QA and SRP approaches need 96.35\% and 97.08\% fewer requests, respectively, compared to the ``All Slots'' approaches, and 86.33\% and 89.06\% fewer requests, respectively, compared to the ``Turn Domains Slots'' approaches. However, compared to a baseline method that includes all schema slots for each turn in a single prompt, our approaches necessitate 120.16\% and 77.02\% more requests, respectively. This increase is due to a significant reduction in prompt length, as we now use only the slots of the active domains rather than all domains. Additionally, there is a substantial improvement in JGA when compared to the results and prompts reported by \citet{heck-etal-2023-chatgpt}.
   \todo[disable,size=\small,inline]{not very clear to me with this much reduction in efficiency, we get this much performance improvement maybe mention small models as well}
    \todo[disable,size=\small,inline, color=yellow]{Done. I rephrased the paragraph. Originally. it  was: However, compared to a baseline method where all schema slots for each turn are included in a single prompt, our approaches require 120.16\% and 77.02\% more requests, respectively. This increase can be attributed to significant impacts on prompt length and model performance when compared to the results and prompts reported by \citet{heck-etal-2023-chatgpt}.}

    \begin{table}[ht]
    \centering
        \scalebox{0.73}{
        \begin{threeparttable}
            \begin{tabular}{lcc}
                \toprule
                \textbf{DST Approach} & \textbf{\# LLM API Requests}\\
                \midrule
                All Slots\tnote{a} & 449.7\\
                Turn Domains' Slots\tnote{b} & 120.2 \\
                All Slots in One Prompt\tnote{c} & 7.4 \\
                DST-as-QA & 16.4\\
                DST-as-SRP & 13.1 \\
                \bottomrule
            \end{tabular}
            \begin{tablenotes}
                \footnotesize
                \item[a] Track all slots from all domains each in a single LLM request
                \item[b] Track slots from active domains and each in a single LLM request
                \item[c] Track slots from all domains but in a single LLM request
            \end{tablenotes}
        \end{threeparttable}
        }
        \caption{Comparison of Different Approaches and Their Number of LLM Prompt Requests for the MultiWOZ}
        \label{tab:llm_queries}
    \end{table}

%% file: sections/conclusion.tex
In this work, we introduce a zero-shot, open-vocabulary Dialogue State Tracking (DST) system that integrates domain classification and DST in a single pipeline. By reformulating DST as a question-answering task and employing adaptable prompting techniques, our system adapts to new slot values without additional fine-tuning. We find that by selecting appropriate techniques---either QA-based or well-structured prompts tailored to the size of the language model---we can surpass SOTA models even without relying on any predefined values from ontologies. Although integrating the domain classification stage reduces the pipeline performance, it is essential for creating a practical system. Finally, we demonstrate the computational efficiency of our techniques by smartly selecting the slots to query and optimizing the prompts to track all slots in one go, rather than querying the language model for each slot individually.

%% file: sections/ethics.tex
The disclaimers for GPT-4-Turo and Gemini state that these models may produce inaccurate information and make mistakes. All models, code, and datasets were used in compliance with their respective licenses, terms of use, and intended purposes. We have provided the code and prompt templates developed for this work. The data we used and generated does not contain any information that names or uniquely identifies individual people, nor does it include offensive content.

%% file: sections/limitations.tex
The performance of the Self-Refined Prompt (SRP) approach is dependent on the specific language model variant employed. Each model, has unique characteristics and capabilities that influence how well it can interpret and execute the prompts. Consequently, the SRP method requires careful tuning and adjustments for each model variant to achieve optimal performance. This process involves iteratively refining prompts. Additionally, the reliance on specific model variants means that updates or changes to these models by their developers could necessitate further adjustments to the SRP approach. 

%% file: sections/appendix.tex
\appendix  
            \twocolumn[{%
            \section{Prompt Templates}
            \label{app:prompt-template}
            \subsection{Domain Classification}
            \label{app:dc_prompt-template}
            
            \textbf{Task Description}
            
            For the dialogue turns between the user and system, Which service is the user asking for? Choose from the following: restaurant, attraction, hotel, taxi, train, bus, hospital, police. Classify the user's turn based on user intents, dialogue context, and previous turns. Don't be biased toward domain-specific terms. If you can't find a service, return None.
            
            \begin{center} 
            \raggedright
            \begin{alltt}
            \textbf{Final Prompt:}
            
            Consider the following domains or services:
            
            restaurant, attraction, hotel, taxi, train, bus, -hospital, police
            
            Now consider the successive turns that I will provide you between two speakers: a USER and a SYSTEM. Which of the domains (one or more domains) the user is asking service for?
            
            \textbf{Guidelines:}
            Follow the following 3 instructions:
            
            - Classify the user's turn based on the intents, context and previouse turns
            
            - If the user's turn involves multiple domains, classify it under all relevant domains.
            - If the user's turn doesn't include a service inquiry, return None.
            
            \textbf{Format the output in json array with 'domains' as key and no more details.}

            \end{alltt}
            \end{center}

    \subsection{Entity Extraction}
        \label{app:entity-extraction-prompt-template}
        \begin{center} 
        \raggedright
        \begin{alltt}
            I will provide you the definition of the entities you need to extract, the sentence from where your extract the entities and the output format.

            \textbf{Entity definition:}
            -TIME: explicit time values. Please normalize the time to 24-format.
            
            -NUMBER: Any format of number.
            
            -PRICE: price
            
            -LOCATION:  geographic location, address, city, town or area
            
            -NAME: Name of hotel, train station, restaurant or attraction
            
            -CODE:  reference number, postcode or id.
            
            -BOOLEAN: true or false for exists or doesn't exist
            
            \textbf{Output Format:} json with the following keys:
            \textbf{\{entities\}}
            If no entities are presented in any categories keep it [].

            \textbf{\{turn\}}
            
            \textbf{Output:} Let's analyze it step-by-step and extract the values carefully. If you are not sure about any value, don't return it. Focus on the value, not the abstract entity.
        \end{alltt}
        \end{center}

    \subsection{Slot Value Multiple Choice Questions}
        \label{app:slot-value-question-prompt-template}
        \begin{center} 
        \raggedright
        \begin{alltt}
            Consider the dialogue below between USER and SYSTEM:
            
            \textbf{\{dialgoue}\}
            
            Can you select the value of the \textbf{\{slotname\}} in the last turn (turn index  \textbf{\{turnindex\}}) from the list below? 
            
            \textbf{\{slotvalues\}}
                        
            \textbf{Guidelines:}

            - Return the answer in JSON with the \textbf{\{slotkey\}} as key.            
            - Don't assume value and just return values from the last turn (turn index \textbf{\{turnindex\}}).
        \end{alltt}
        \end{center}
    }]
        \twocolumn[{%
        \subsection{Dialogue State Tracking}
        \label{app:dst-prompt-template}
        \textbf{Task Description}

        Consider the following slots and their definitions: \{slots\}. I will show you a conversation between a USER and a SYSTEM about \{domain\}. Your job is to find and note the slot values mentioned in each turn. If a value is given by the speaker, write it down. If the speaker accepts a value from the previous turn, include that too. If the speaker asks about a slot, mark it as "?". If the speaker says they don’t have a preference, mark it as "*". Only include slots that are mentioned, and return the slot values as a JSON object with these keys: \{slotnames\}.

        \textbf{GPT-4-Turbo}
        \begin{center} 
        \raggedright
        \begin{alltt}
            Consider the list of concepts, called "slots", provided below with their definitions:

            \textbf{\{slots\}}
            
            Now consider the successive turns that I will provide you between two speakers: a USER and a SYSTEM about \text{\{domain}\}. Please meticulously extract and catalog the slot values from each pairs of turns based on the provided slot definitions and follow the following 6 instructions
            
            1. Carefully identify the slot values explicitly mentioned by the speaker in that turn.
            
            2. Ensure you incorporate any acknowledged or accepted slot values from the directly previous turn within the current speaker's turn.
            
            3. For any direct inquiry by the speaker about a specific slot, mark its value as "?".
            
            4. Carefully identify the slots being asked about by the speaker and mark their values as "?".
            
            5. If the speaker explicitly mentions they have no preference or it doesn't matter for a specific slot, mark its value as "*".
            
            6. If a slot isn't mentioned in a turn, do not include it.

            Ensure thoroughness and accuracy in the identification process. Return the output as json object with the following as key and their values and no more details:\textbf{\{slotsnames\}}
        \end{alltt}
        \end{center}

        \textbf{GPT-3.5-Turbo}
        \begin{center} 
        \raggedright
        \begin{alltt}
                As a dialogue state tracker, your task is to track the slot values that are important to the user during a series of dialogue turns between a USER and a SYSTEM. We are interested in capturing the user's preferences and inquiries about \textbf{\{domain\}} regarding specific slots.
                Slots to Track:
                \textbf{\{slots\}}
    
                \textbf{Instructions:}
                1. Track slot values mentioned by the user during each dialogue turn.
                
                2. If the system mentions relevant slot values that are important to the user's context or preferences, track those as well.
                
                3. If the user explicitly states they have no preference or don't care about a specific slot, set its value to *
                
                4. Provide the slot values in a JSON format.
                
                5. Make sure to check all the slots, and don't miss any.

                \textbf{Output Format: json object}
                
                \textbf{\{
                  slotname: slotvalue
                }\}
                
                With the following key: \textbf{\{slots\}}
        \end{alltt}
        \end{center}
        }]
        \twocolumn[{%
        \textbf{Llama 3}
        \begin{center} 
        \raggedright
        \begin{alltt}
                As a dialogue state tracker, your task is to track the following {domain} slots during the dialogue turns that I will provide afterwards:
                slots:
                  \textbf{\{slots\}}
                
                \textbf{Instructions:}
                
                1- Slot actual value: if the user mentioned the slot value
                
                2- Slot actual value: if the system mentioned the slot value and the user didn\'t reject
                3- *: if the user states he has no preference
                4- Sometimes you need to get the slot value from another domain if the user refers to it.
                5- Otherwise, don\'t return it.
                
                \textbf{-OutputFormat: json object}
                \textbf{\{slotname: slotvalue \}}
                With the following key: \textbf{\{slots\}}

        \end{alltt}
        \end{center}
        \vspace{0.3cm}
        }]
\section{Language Models Hyperparameters}
\label{app:llms-parameters}
    Table \ref{tab:llm-parameters} presents the values of the temperature and top-p parameters used for different models in the domain classification and DST tasks.
    \begin{table}[htbp]
        \centering
        \begin{tabular}{lcc}
            \toprule
            \textbf{Model} & \textbf{Temperature} & \textbf{Top-p} \\
            \midrule
            \multicolumn{3}{c}{\textbf{Domain Classification Parameters}} \\
            \midrule
            GPT-4-Turbo & 0.3 & 0.9 \\
            Gemini 1.0 & 0.8 & 1 \\
            Llama 3 & 0.25 & 0.9 \\
            Qwen 1.5 & 0.25 & 1 \\
            Mixtral v0.1 & 0.25 & 0.9 \\
            \midrule
            \multicolumn{3}{c}{\textbf{Dialogue State Tracking Parameters}} \\
            \midrule
            GPT-4-Turbo & 0.5 & 0.9 \\
            Gemini 1.0 & 0.9 & 1 \\
            Llama 3 & 0.7 & 0.9 \\
            Mixtral v0.1 & 0.25 & 1 \\
            Qwen 1.5 & 0.25 & 1 \\
            \bottomrule
        \end{tabular}
        \caption{Temperature and Top-p Parameters for the LLMs Models in Domain Classification and DST}
        \label{tab:llm-parameters}
    \end{table}

\section{LLM Request-Based Metrics for Resource Efficiency}
\label{app:llm-requests}
    We chose the number of prompt requests as a primary metric because it offers a consistent and practical means to evaluate resource efficiency across LLMs. Both time and cost are tightly linked to the frequency of API calls, particularly when using proprietary models where each request has an associated cost and latency. By focusing on the number of requests, we can bypass the complexities of traditional time complexity analysis, which is challenging to apply here given the black-box nature of these APIs. The lack of insight into system-level performance and hardware makes it impractical to assess time complexity in conventional terms, so tracking API requests provides a more straightforward and reliable measure of efficiency.

    While prompt length is also a factor, it introduces additional variability that complicates analysis. Prompt length can depend on task details, slot values, descriptions, and dialogue structure, as well as API-specific requirements—some LLMs, for instance, necessitate resending the entire conversation history with each new prompt, while others do not. These differences further underscore the value of using the number of prompt requests as a central metric, allowing us to account for efficiency in a way that remains consistent regardless of such model-specific nuances.
    
    The following outlines the different scenarios for prompting a LLM based on the availability of domain information, along with the corresponding calculations for the number of the required prompt requests:
     \begin{itemize}
      \item All Slots: Query the LLM about every slot in the dataset. The number of LLM queries is equal to the total number of slots in the dataset ontology multiplied by the total number of turns in the test split
    
      \item Domain Slots: Query the LLM about each slot in that domain. The average number of queries is equal to the average number of domains per turn multiplied by the average number of slots per domain for each turn. Summing these values for all turns in the test split gives the total
    
      \item In SRP, you need to prompt the LLM once per domain (turn domains that were determined in the previous step) per turn. The total number of LLM queries equals the average number of domains per turn multiplied by the total number of turns in the test split.
    \end{itemize}

\section{Domain Classification Accuracy Rate per Turn in MultiWOZ 2.4}
    \label{app:domain-classification-accuracy}
\begin{figure}[htbp]
                \centering
                \includegraphics[width=0.5\textwidth]{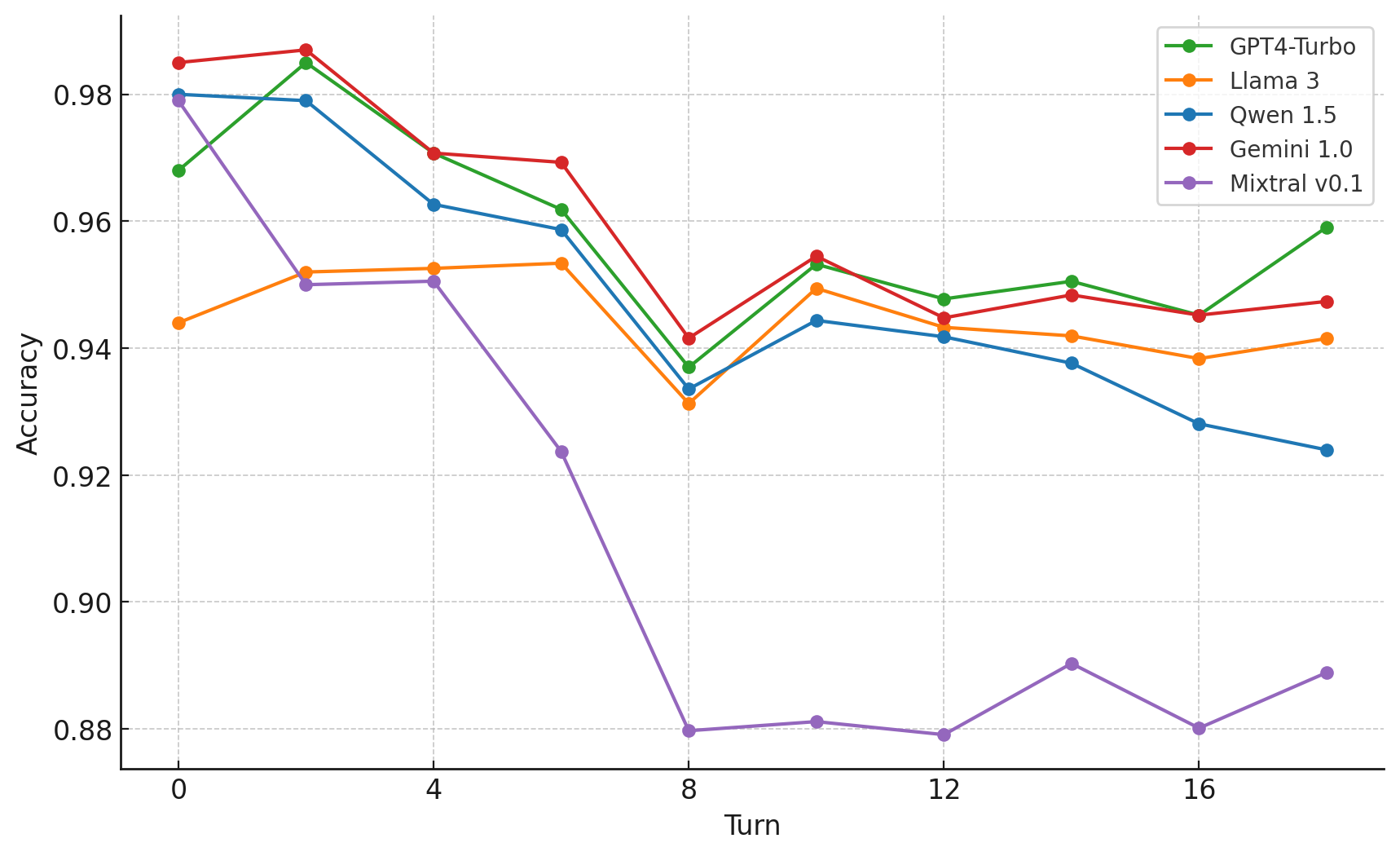}
                \caption{Turn Domain Classification Accuracy for MultiWOZ 2.4}
                \label{fig:turn_domain_class_acc}
        \end{figure}

\begin{figure}[htbp]
            \centering
            \includegraphics[width=0.5\textwidth]{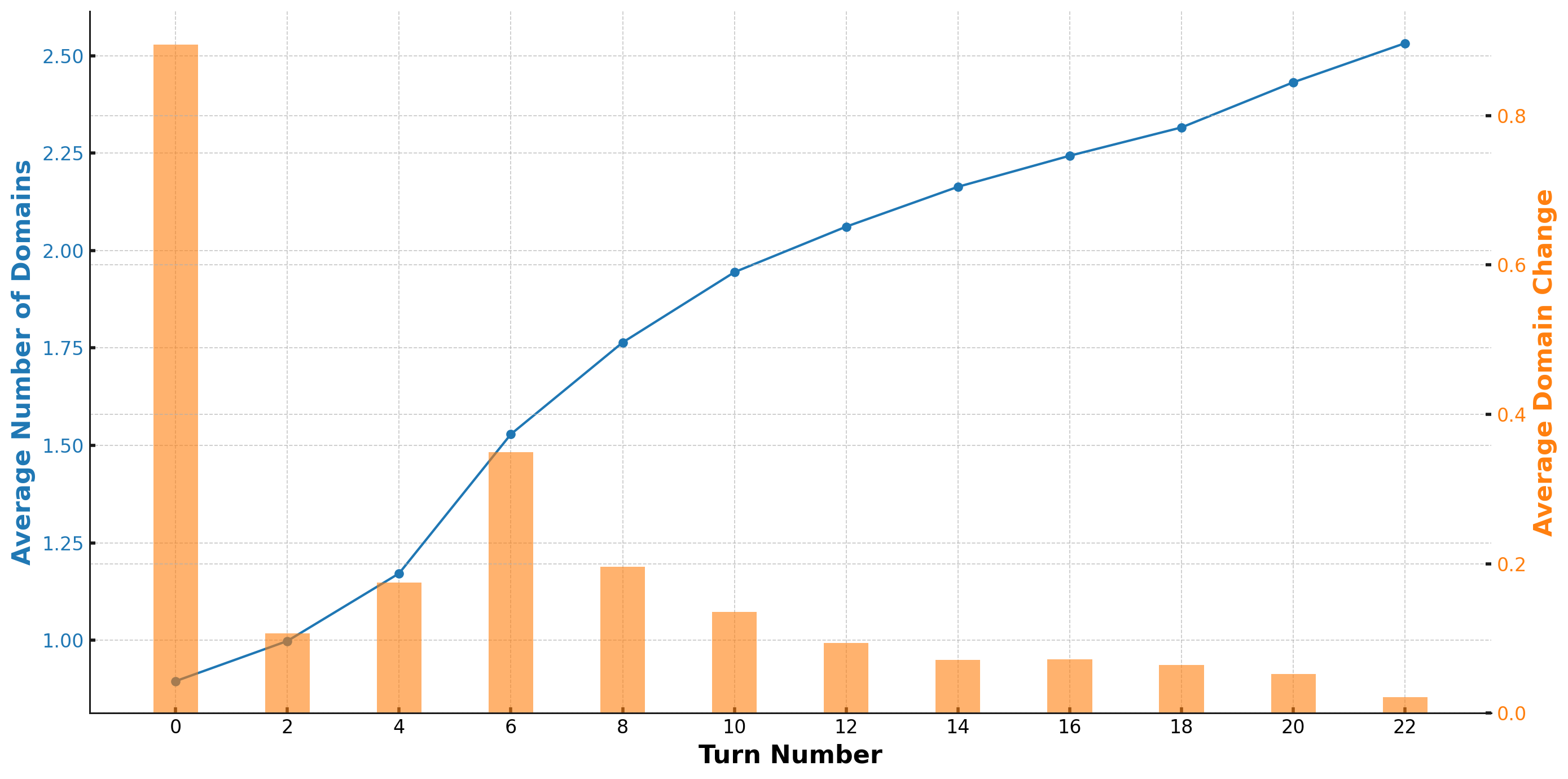}
            \caption{Average Number of Domains and Domain Change Per Turn for MultiWOZ 2.4}
            \label{fig:mwoz-domain-change}
        \end{figure}

    We analyze the performance of the Domain Classification models w.r.t turn numbers to gain insights into model behavior over the progression of dialogues. Fig.~\ref{fig:turn_domain_class_acc} shows the domain classification accuracy per turn in MultiWoz 2.4. All models generally show an increasing trend in classification accuracy as the dialogue progresses, suggesting improved performance with richer context. However, there is a notable dip in accuracy around Turn 8 across all models, indicating a common challenge in handling dialogue complexity at this stage. After a deeper analysis, we find the reason to be the domain change rates, i.e., how many times the domain changes at a particular turn.
    Following the domain classification accuracy results, we find that Gemini and GPT4-Turbo providing the best performance, independent from the turn number.
    The average number of domain changes per turn is determined by identifying the new domains introduced at each turn. For each dialogue in the test split of the datasets, we iterate through all the turns, noting the new domains introduced at each one. We then sum the number of new domains for each turn and divide this sum by the total number of dialogues that include that specific turn number. Similarly, the average number of domains in each turn is found by summing up the number of active domains up to that turn, and then dividing this sum by the number of dialogues that contain that specific turn number. Figure \ref{fig:mwoz-domain-change} shows these metrics per turn.

\section{SGD Domain Classification Results Analysis}
\label{app:sgd-cd-analysis}
\begin{figure}[htbp]
            \centering
            \includegraphics[width=0.5\textwidth]{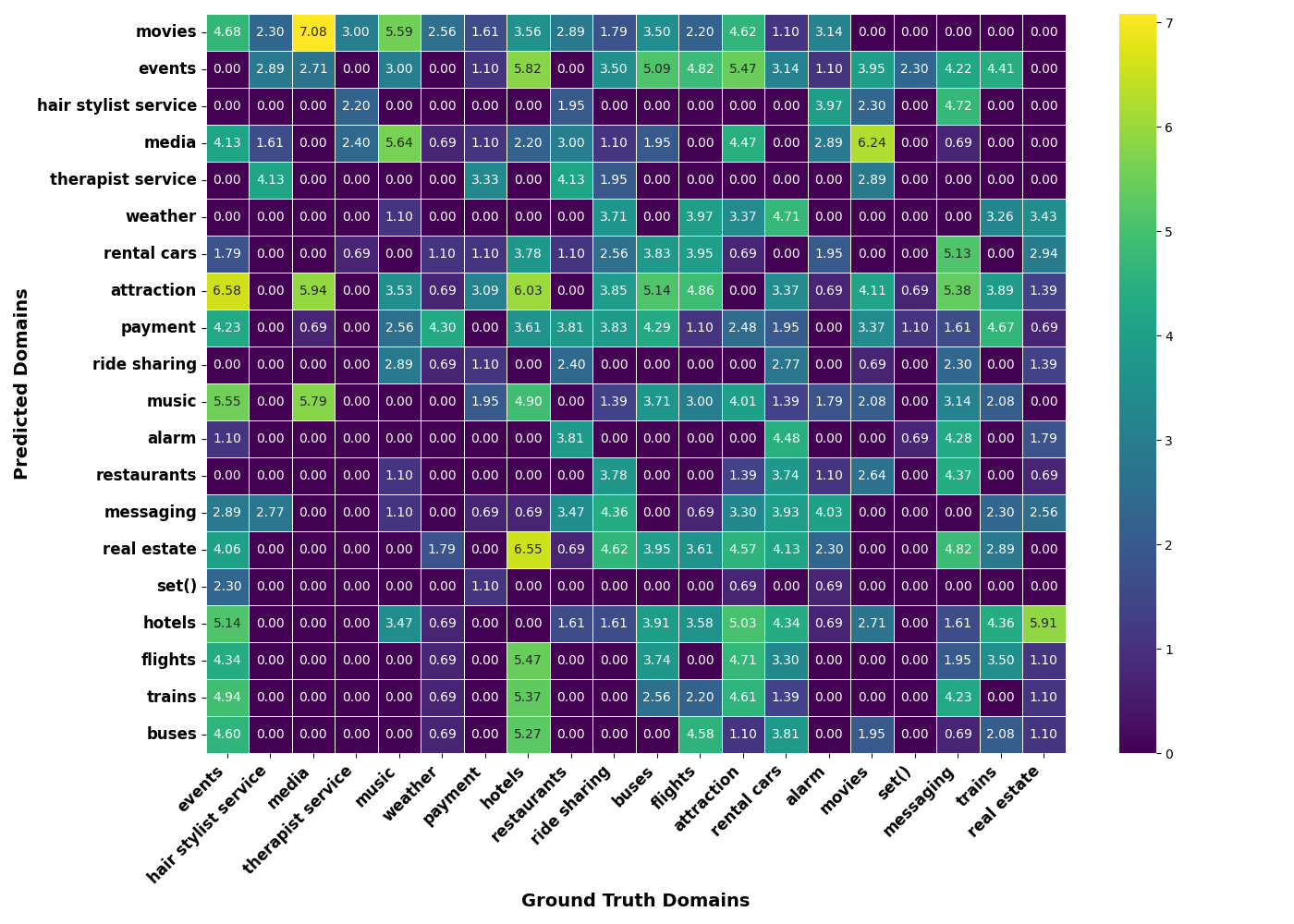}
            \caption{Ground-Truth Domains vs Incorrectly Predicted Ones in SGD}
            \label{fig:sgd-misclassified-domains-heatmap}
    \end{figure}

    Referring to Table \ref{tab:domain_class}, both GPT-4-Turbo and Llama experienced a drop in domain classification accuracy for the SGD dataset compared to the MultiWOZ dataset. We visualized the domain misclassifications using a heatmap of predicted versus ground truth domains, revealing key areas of confusion in figure \ref{fig:sgd-misclassified-domains-heatmap}.
    The heatmap of domain misclassifications reveals significant patterns where the model struggles to differentiate between similar contexts, as described in the dataset ontology. Media is frequently misclassified as movies, likely due to the overlap in content descriptions. The media domain encompasses a wide range of content, including shows, movies, sports, and documentaries, which overlaps significantly with the movies domain focused on searching for showtimes and booking movie tickets. Similarly, movies are often mistaken for events because both domains share terminology related to showtimes and venues, given that movies often involve events like premieres and screenings.

The attraction domain is confused with both events and hotels. Attractions, described as tourist spots and points of interest, often host events and are linked with nearby accommodations, leading to misclassifications. For example, an attraction might host a special event or require hotel stays, creating overlapping contexts. Misclassifications also occur between hair stylist service and therapist service, both of which involve personal care appointments. The hair stylist service domain involves finding and reserving hair stylists and salons, while therapist service involves finding and reserving therapists, with both domains sharing similar appointment-related vocabulary.

The heatmap further shows that hotels are often misclassified as events, reflecting the dual role of hotels as event venues. Hotels frequently host conferences, weddings, and other events, which explains the shared terminology and resulting confusion. Additionally, confusion between ride sharing and restaurants likely stems from discussions about transportation to dining locations. The ride sharing domain, which focuses on booking cabs, overlaps with the restaurants domain where transportation to dining venues is commonly discussed. These findings highlight the need for more detailed descriptions that accurately describe the domains services, particularly for domains with overlapping contexts as described in the ontology.

\section{SRP Performance Per Slot}
\label{app:mwz-srp-slot-jga}
\begin{figure*}[htbp]
                \centering
                \includegraphics[width=\textwidth]{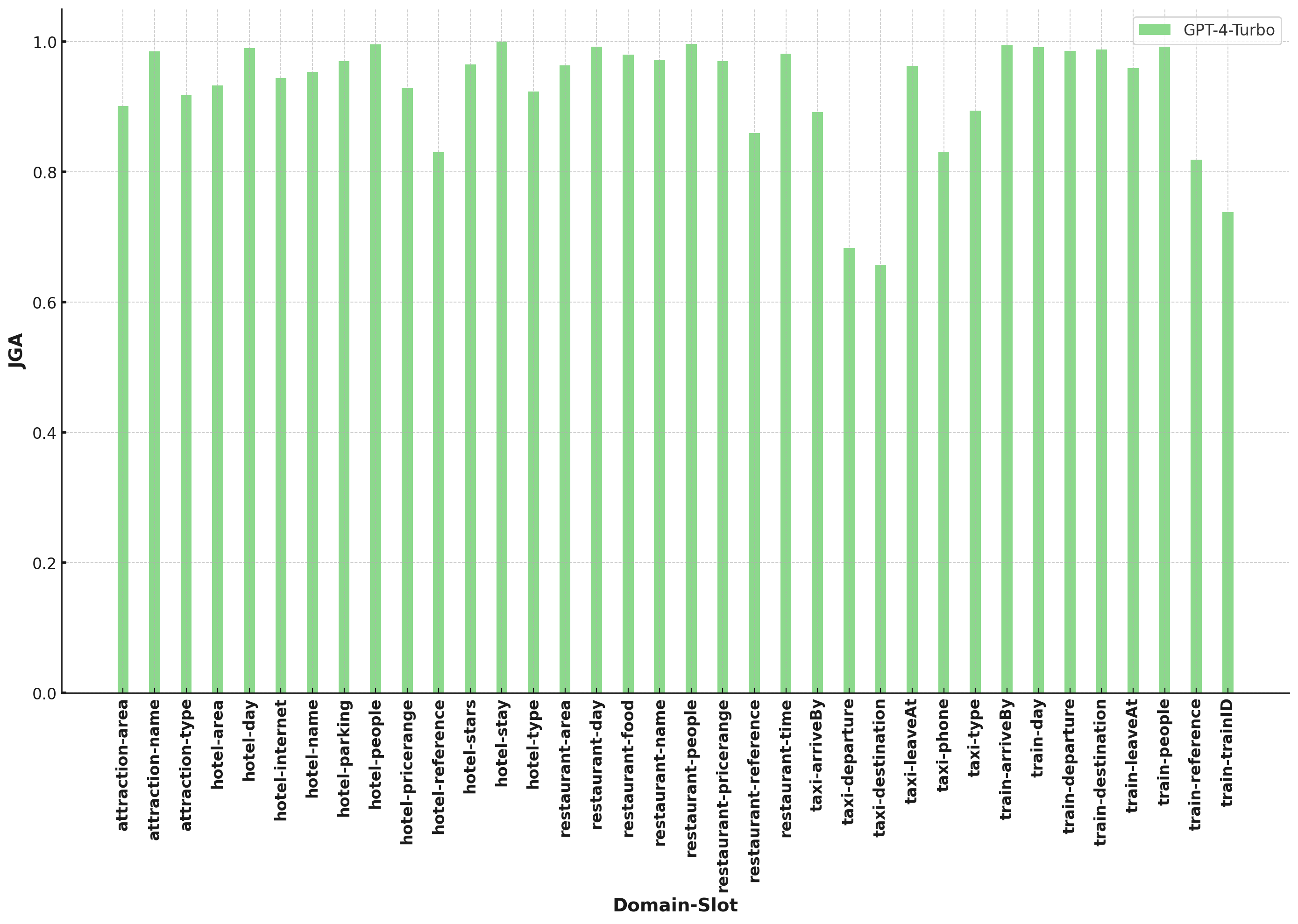}
                \caption{Slot Value Accuracy for the MultiWOZ 2.4 Using GPT-4-Turbo}
                \label{fig:mwz-slot-accuracy-gtp-4-turbo}
        \end{figure*}

\begin{figure*}[ht]
                \centering
                \includegraphics[width=\textwidth]{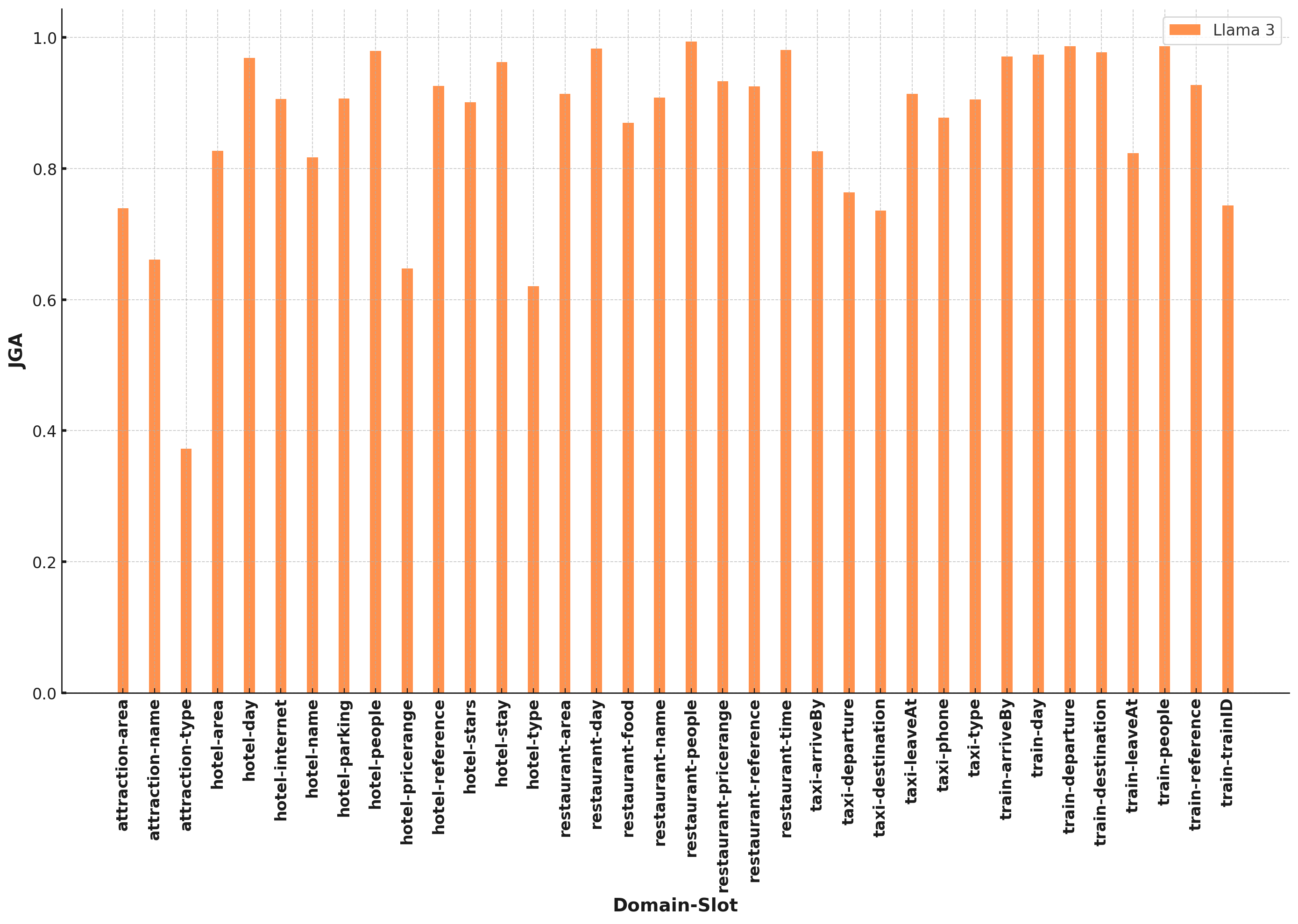}
                \caption{Slot Value Accuracy for the MultiWOZ 2.4 Using Llama 3}
                \label{fig:mwz-slot-accuracy-llama3}
        \end{figure*}

    In addition to analyzing the SRP performance per domain, we visualize its performance per slot in Fig.\ref{fig:mwz-slot-accuracy-gtp-4-turbo} and \ref{fig:mwz-slot-accuracy-llama3}

\section{Entities-Slots Mapping}
\label{app:entity-slots}
Table \ref{tab:slot-entity-map} below shows the entity-type to slot map for MultiWOZ dataset.

    \begin{table*}[htbp]
        \centering
        \begin{tabular}{|c|l|}
            \hline
            \textbf{Entity Type} & \textbf{Slots} \\
            \hline
            NAME & Restaurant.name, Attraction.name, Hotel.name, Taxi.destination, \\
                 & Taxi.departure, Train.train-id, Hospital.department, Police.name \\
            \hline
            DAY & Restaurant.book-day, Hotel.book-day, Train.day, Bus.day \\
            \hline
            TIME & Restaurant.book-time, Attraction.open-hours, Taxi.leave-at, Taxi.arrive-by, \\
                 & Train.leave-at, Train.arrive-by, Bus.leave-at \\
            \hline
            LOCATION & Restaurant.area, Restaurant.address, Attraction.area, Attraction.address, \\
                     & Hotel.area, Hotel.address, Train.destination, Train.departure, \\
                     & Bus.destination, Bus.departure, Hospital.address, Police.address \\
            \hline
            NUMBER & Restaurant.book-people-count, Restaurant.phone, Hotel.book-people, \\
                   & Hotel.book-stay, Hotel.phone, Hotel.stars, Train.duration, \\
                   & Train.book-people-count, Taxi.phone, Attraction.phone, \\
                   & Hospital.phone, Police.phone \\
            \hline
            PRICE & Attraction.entrance-fee, Train.price \\
            \hline
            TYPE & Restaurant.food, Attraction.type, Hotel.type, Taxi.type \\
            \hline
            RANGE & Restaurant.price-range, Hotel.price-range \\
            \hline
            DONTCARE & All \\
            \hline
            CODE & Restaurant.post-code, Restaurant.reference-code, Attraction.post-code, \\
                 & Hotel.post-code, Hotel.reference-code, Train.reference-code, Police.post-code \\
            \hline
            BOOLEAN & Hotel.parking, Hotel.internet \\
            \hline
        \end{tabular}
        \caption{Entity Type-Slot Mapping}
        \label{tab:slot-entity-map}
    \end{table*}